\documentclass[a4paper,11pt,twocolumn,twoside]{article}
\usepackage[dvips]{graphicx}
\usepackage{sepln_en}
\usepackage{fullname}
\usepackage[utf8]{inputenc}
\usepackage{float}
\usepackage{tabularx}
\usepackage{subcaption}
\usepackage{booktabs}
\usepackage{xltabular}
\usepackage{xurl}
\usepackage{todonotes}

\input epsf

\title{Lessons learned from the evaluation of Spanish Language Models}

\author {\textbf{Rodrigo Agerri}, \textbf{Eneko Agirre} \\
HiTZ Center - Ixa, University of the Basque Country UPV/EHU\\
rodrigo.agerri@ehu.eus, e.agirre@ehu.eus\\
}

\seplntranstitle{Conclusiones de la evaluación de
Modelos del Lenguaje en Español}

\seplnclave{Modelos de Lenguaje, Clasificación de Textos, Etiquetado Secuencial, Procesamiento del Lenguaje Natural.}

\seplnresumen{Actualmente existen varios modelos del lenguaje en español (también conocidos como
	BERTs) los cuales han sido
	desarrollados tanto en el marco de grandes proyectos que utilizan corpus
	privados de gran tamaño, como mediante esfuerzos académicos de menor escala aprovechando datos de libre acceso.
	En este artículo presentamos una comparación exhaustiva de modelos de
	lenguaje en español con los siguientes resultados: (i) La inclusión de modelos multilingües previamente ignorados
	altera sustancialmente el panorama de la evaluación para el español, ya que
	resultan ser en general mejores que sus homólogos monolingües;
	(ii) Las diferencias en los resultados entre los modelos monolingües no son concluyentes, ya que
	aquellos supuestamente más pequeños e inferiores obtienen resultados más
	que competitivos. El resultado de nuestra evaluación demuestra que es
	necesario seguir investigando para comprender los factores que subyacen a estos resultados.
	En este sentido, es necesario seguir investigando el efecto del tamaño del corpus, su calidad y las técnicas de preentrenamiento
	para poder obtener modelos monolingües en español significativamente mejores que los multilingües
	ya existentes. Aunque esta actividad reciente demuestra un creciente interés en el desarrollo
	de la tecnología lingüística para el español, nuestros resultados ponen de
	manifiesto que el desarrollo de modelos de lenguaje sigue siendo un
	problema abierto que requiere conjugar recursos (monetarios y/o computacionales) con los mejores conocimientos y 
	prácticas de investigación en PLN.}

\seplnkey{Masked Language Models, Text Classification,
Sequence Labelling, Natural Language Processing.}

\seplnabstract{Given the impact of language models on the field of Natural
	Language Processing, a number of Spanish encoder-only masked language
	models (aka BERTs) have been trained and released. These models were
	developed either within large projects using very large private corpora or
	by means of smaller scale academic efforts leveraging freely available data. In
	this paper we present a comprehensive head-to-head comparison of language
	models for Spanish with the following results: (i) Previously ignored multilingual models
	from large companies fare better than monolingual models, substantially changing the evaluation landscape of language models in
	Spanish;
	(ii) Results across the monolingual models are not conclusive, with
	supposedly smaller and inferior models performing competitively. Based on
	these empirical results, we argue for the need of more research to understand
	the factors underlying them. In this sense, the effect of corpus size, quality and pre-training
	techniques need to be further investigated to be able to obtain Spanish
	monolingual models significantly better than the multilingual ones released
	by large private companies, specially in the face of rapid ongoing progress 
	in the field. The recent activity in the development
	of language technology for Spanish is to be welcomed, but our results show that building language 
	models remains an open, resource-heavy problem which requires to marry resources 
	(monetary and/or computational) with the best research expertise and practice.
}

\firstpageno{1}

\begin{document}


\setlength\titlebox{25cm} 

\label{firstpage} \maketitle

%

\section{Introduction}

Deep Learning has changed the application and research landscape in Natural
Language Processing (NLP). The field has experienced a paradigm shift that has
rendered previous techniques obsolete for many tasks, and nowadays large
companies such as Google or Meta rely on deep learning techniques to develop
NLP applications. Central to these developments lay large pre-trained language
models, which are trained on gigantic corpora (e.g. crawls of the entire Web)
requiring costly hardware. The cost of developing and training such models is
so high that most recent innovations come from such large companies and focus
on English. Thus, the best available language models for English have been released
to the public by large companies. Furthermore, in some cases large language
models that are currently being used are not even released, but offered instead
as a pay-per-use API.

A natural question arises regarding languages other than English, as the same large
companies have published multilingual versions of these models with support for 100
languages, such as multilingual BERT and XLM-RoBERTa
\cite{Devlin19,Conneau2020UnsupervisedCR}. While these multilingual models excel in many NLP tasks involving
high-resourced languages such as English, their performance is not always as
good as monolingual models. In fact, recent studies seem to suggest that a
careful training design and appropriate corpora selection results in better
models for each specific language
\cite{martin-etal-2020-camembert,agerri-etal-2020-give,Agerri2020ProjectingHA}.
Although several language model architectures exist, most efforts building
monolingual models have focused on encoder-only masked language
models (e.g. BERT and variants) \cite{Devlin19,Liu2019RoBERTaAR}, so we will leave decoder-only
causal language models (e.g. GPT) and encoder-decoder models (e.g. T5) for
future analysis \cite{gpt3,opt,bloom,t5,xue-etal-2021-mt5}.

Thus, following previous work comparing monolingual and multilingual models
\cite{Vries2019BERTjeAD,Virtanen2019MultilingualIN,martin-etal-2020-camembert,Agerri2020ProjectingHA,Tanvir2021EstBERTAP,armengol-estape-etal-2021-multilingual},
in this paper we are going to focus on Spanish, for which several encoder-only
masked language models  have been trained and released
\cite{CaneteCFP2020,gutierrezfandino2022,BERTIN}. The models have been
developed either in heavily-subsidized projects with very large corpora or in
smaller scale academic efforts on more limited, freely available corpora. In
order to compare the quality of the language models, we follow usual practice
and perform a downstream evaluation where all language models are treated
equally and applied to a large set of Spanish NLP evaluation datasets,
including common tasks such as part-of-speech tagging, named-entity reconition,
natural language inference, semantic textual similarity, question answering,
paraphrasis or metaphor detection. However, unlike previous evaluations for
Spanish, we do include in our evaluation widely used multilingual models such
as XLM-RoBERTa and mDeBERTa \cite{Conneau2020UnsupervisedCR,He2021DeBERTaV3ID}.

Our comprehensive head-to-head comparison yields surprising results: 
(i) Considering the previously ignored XLM-RoBERTa and mDeBERTa 
substantially change the evaluation landscape of language models in
Spanish, as they happen to fare better than their monolingual counterparts.
In particular, our results show that XLM-RoBERTa-large, released by
Meta in 2020 \cite{Conneau2020UnsupervisedCR} obtains the best results in
the majority of the tasks. Furthermore, mDeBERTa \cite{He2021DeBERTaV3ID}, 
a smaller base-size model, performs second overall. (ii) Despite claims to the
contrary \cite{gutierrezfandino2022}, results among the monolingual models
are quite close, and supposedly smaller and inferior models such as
IXABERTesv2\footnote{\url{http://www.deeptext.eus/eu/node/3}} obtaining similar
or better results with respect to the the MarIA RoBERTa-bne models; (iii)
In addition to downstream evaluation, the effect of corpus size,
corpus quality and pre-training techniques need to be further investigated
\cite{martin-etal-2020-camembert,Artetxe2022DoesCQ} to advance current state-of-the-art in language
models; (iv) despite the strong results obtained by evaluating the language
models, for some tasks they remain well below the state-of-the-art. Code and
data is publicly available to facilitate research on this topic and
reproducibility of results\footnote{\url{https://github.com/ragerri/evaluation-spanish-language-models}}.

Based on this findings, we argue for more research to understand the
factors underlying the results and to be able
to obtain Spanish monolingual models significantly better than the multilingual
ones released by large private companies. While this recent activity building models bodes well the
development of language technology for Spanish, our results show that building
language models remains an open, resource-heavy problem which requires to marry
resources (monetary and/or computational) with the best research expertise and
practice.

The rest of the paper is structured as follows. Next section discusses related work on
monolingual and multilingual language models. Section
\ref{sec:spanish-language-models} provides details of the language models for
Spanish that will benchmarked in Section \ref{sec:results} following the
experimental setup of Section \ref{sec:experimental-setup}. In Section
\ref{sec:discussion} we will go over the lessons learned quite throughly and we
will finish with some concluding remarks.

\section{Related Work}\label{sec:related-work}

The release of encoder-based masked language models (MLMs) for English caused a
paradigm-shift in Natural Language Processing (NLP) research. After the
original BERT model \cite{Devlin19}, many variations and improvements were
quickly developed \cite{Liu2019RoBERTaAR,He2021DeBERTaV3ID}. At the same time,
large multilingual models such as multilingual BERT and XLM-RoBERTa, trained to work on 100 languages, were published,
with extraordinary results both monolingual and, especially, on multilingual
and cross-lingual settings
\cite{Pires2019HowMI,Wu2020AreAL,Conneau2020UnsupervisedCR}. The availability
of such multiligual models posed the question whether they were the optimal
solution for other languages different to English. This in turn caused the
appearance of a large body of research studying the performance of such
multilingual models on specific languages, often in comparison to monolingual
counterparts specifically tailored to the target language
\cite{Nozza2020WhatT}. 

Recent studies suggest that while the multilingual models excel in many NLP
tasks involving high-resourced languages such as English, their performance is
not usually as good as monolingual models. Thus, previous work on monolingual
models for languages such as Basque or French suggest that a careful training design and appropriate corpora selection results in
better models for each specific language \cite{martin-etal-2020-camembert,agerri-etal-2020-give}.

Other studies focused on the quality of the corpus itself
\cite{Virtanen2019MultilingualIN,Tanvir2021EstBERTAP} while for other languages
such as Basque or Catalan, in addition to
developing language models, a large effort on generating new datasets for
benchmarking was also put in place
\cite{armengol-estape-etal-2021-multilingual,urbizu-etal-2022-basqueglue}.
Finally, recent research has empirically demonstrated that, while size is
important, carefully studying the pre-training method and auditing the quality
of the corpus is crucial to understand the performance of language models on
downstream tasks \cite{kreutzer-etal-2022-quality,Artetxe2022DoesCQ}.

In any case, most of the previous work shows that monolingual models perform in
general better than the multilingual ones, also with respect to XLM-RoBERTa
\cite{martin-etal-2020-camembert,armengol-estape-etal-2021-multilingual}.
However, for Spanish the situation is slightly different because the largest evaluation of
language models for Spanish does not include XLM-RoBERTa or the more recent
mDeBERTa \cite{gutierrezfandino2022}. In this work we will
address this issue by including them in the evaluation of language models
for Spanish.

\begin{table*}[t]
\centering
\begin{tabular}{l|p{2.2cm}rccccc} 
\toprule
	Model & corpus & \#words & L & H & A & V & \#params \\ \midrule
	Multilingual BERT & Wiki & 0.7B & 12 & 768 & 12 & 110K & 110M \\
	BETO & Opus,Wiki & 3B & 12 & 768 & 12 & 30K & 110M \\
	IXABERTesv1 & Gigaword,Wiki & 5.7B & 12 & 768 & 12 & 50K & 110M \\
	ixambert & Wiki & 0.7B & 12 & 768 & 12 & 119K & 110M \\
	IXABERTesv2 & OSCAR & 25B & 12 & 768 & 12 & 50K & 125M \\
	XLM-RoBERTa-base & CC-100 & 9.3B & 12 & 768 & 12 & 250K & 270M \\
	XLM-RoBERTa-large & CC-100 & 9.3B & 24 & 1024 & 16 & 250K & 550M \\
	Electricidad & Opus,Wiki & 3B & 12 & 768 & 12 & 31K & 110M \\
	BERTIN & mC4-es & $~$47B & 12 & 768 & 12 & 50K & 125M \\
	RoBERTa-base-bne & BNE & 135B & 12 & 768 & 12 & 50K & 125M \\
	RoBERTa-large-bne & BNE  & 135B & 24 & 1024 & 16 & 50K & 350M \\
	mDeBERTa & CC-100 & 9.3B & 12 & 768 & 12 & 250K & 198M \\
\bottomrule
\end{tabular}
\caption{Spanish Language Models (in approximate order of creation). L: layer
size; H: hidden size; A: attention heads; V: vocabulary.}  
\label{tab:spanish-lm}
\end{table*}

\section{Spanish Language models}\label{sec:spanish-language-models}

Spanish has been quite a newcomer in the Transformer-based language model
fever, which was hard to understand given that Spanish is the fourth most spoken
language in the world. Thus, while the
number of language-specific models proliferated at a vertiginous rhythm for
many world languages, BETO \cite{CaneteCFP2020} remained the only language
model for a surprisingly large period of time. BETO follows a BERT-base architecture 
and was released around the end of 2019 by researchers at the University of
Chile\footnote{\url{https://github.com/dccuchile/beto}}. The model was trained
on a collection of corpora which included the Spanish Wikipedia and the OPUS
Spanish corpus \cite{Tiedemann2020OPUSMTB} and it was evaluated on the GLUES (short
for GLUE in Spanish)
dataset\footnote{\url{https://github.com/dccuchile/glues}}, comparing
favourably with respect to multilingual BERT.

However, once started, language models for Spanish quickly proliferated. 
In 2020 two models, based on BERT and RoBERTa-base (IXABERTesv1 and v2), were
released\footnote{\url{http://www.deeptext.eus/eu/node/3}} by the Ixa Group of
the University of the Basque Country. This group also published that year 
a multilingual model for Basque, Spanish and English, ixambert,
following the BERT-base architecture \cite{Otegi2020ConversationalQA}.

One year later, a community-based effort coordinated within the Flax/Jack
Community Week organized by HuggingFace
released
BERTIN\footnote{\url{https://huggingface.co/bertin-project/bertin-roberta-base-spanish}}
a RoBERTa-base model \cite{BERTIN}. This model was trained on the
Spanish portion of the mC4 dataset \cite{xue-etal-2021-mt5}. Some of the BERTIN
developers also released an Electra-base Spanish model:
Electricidad\footnote{\url{https://huggingface.co/mrm8488/electricidad-base-discriminator}}. 

Concurrently, a team from the Barcelona Supercomputing Center funded by the
Spanish Government released under the MarIA project\footnote{\url{https://github.com/PlanTL-GOB-ES/lm-
spanish}} two models, RoBERTa-base-bne
and RoBERTa-large-bne, trained on a large corpus based on crawling data from
the Spanish National Library (BNE corpus). The MarIA models were
compared with respect to BETO, BERTIN, Electricidad and multilingual BERT \cite{gutierrezfandino2022}.
Results from other commonly-used multilingual models such as XLM-RoBERTa (both base and
large) or mDeBERTa were not included in the evaluation.

All language models have been trained on publicly available corpora, except the
BNE corpus\footnote{In the paper the MarIA authors mention that it will be
released soon, although at the time of writing the corpus is not available.}.
Public availability is important, as
many features and biases of the language models depend on the corpora where
they have been trained. Furthermore, public availability is required to guarantee reproducibility of
results. It also allows researchers, companies and users to
examine those corpora and thus assess the impact that the features of the
corpora will have in their research and products.

\subsection{Models details}\label{sec:model-details}

Table \ref{tab:spanish-lm} shows the most important details of the language
models we will use in our study, including the corpus type and size on which
they were trained, and technical pre-training details such as the number of
layers, the hidden size, number of attention heads, the vocabulary and the
number of parameters. In the rest of this section we will comment other
relevant aspects to interpret the results reported in Section
\ref{sec:results}. 

BETO, IXABERTesv1 and ixambert are BERT-base models pre-trained with both
Masked Language Modeling (MLM) and Next Sentence Prediction (NSP)
\cite{Devlin19}. BETO performed 2M steps in two different stages: 900K steps
with a batch size of 2048 and maximum sequence length of 128, and the rest of
the training with batch size of 256 and maximum sequence length of 512.
Both IXABERTesv1 and ixambert were trained by executing 1M steps with 256 of batch size and
512 sequence length.

The language models using RoBERTa-base (IXABERTesv2, BERTIN and
RoBERTa-base-bne) and large (RoBERTa-large-bne) are based on the BERT
architecture but (i) trained only on the MLM task, (ii) on larger batches (iii)
on longer sequences and (iv), with dynamic mask generation. While IXABERTesv2 performed 120.500 steps with 2048 batch
size and sequence length 512, BERTIN was trained on 250K steps divided in two
steps: 230k steps with sequences of length 128 and batch size 2048, and the
rest of the training with 512 sequence length and 384 of batch size. Thus, both IXABERTesv2 and
BERTIN roughly follow the RoBERTa approach to pre-training
\cite{Liu2019RoBERTaAR}. However, the MarIA models opted instead for  a batch
of 2048 and 512 sequence length, but reducing the training to one epoch only
with no dropout \cite{Komatsuzaki2019OneEI}.

With respect to the multilingual models, multilingual BERT was trained with a batch
size of 256 and 512 sequence length for 1M steps, using both the MLM and NSP
tasks. Regarding XLM-RoBERTa, both versions were trained over 1.5M steps with
batch 8192 and sequences of 512 length. Finally, mDeBERTa \cite{He2021DeBERTaV3ID} is based on RoBERTa but incorporating
disentangled attention, gradient-disentagled embedding sharing and, most importantly, replacing the MLM task with
replaced token detection (RTD), originally proposed by ELECTRA
\cite{Clark2020ELECTRAPT}; mDeBERTa was trained following the XLM-RoBERTa
procedure but reducing the steps from 1.5M to 500K. 

Thus, the specific pre-training details and the corpora used to generate the
language models substantially differ across the monolingual and the
multilingual models. However, as we will see in the next section, the
fine-tuning performed to evaluate the models on downstream tasks will follow
the same methodology.

\begin{table*}
\begin{center}
\resizebox{1\linewidth}{!}{
\begin{tabular}[b]{l|ccccc|cccc|cc}
\toprule
& \multicolumn{5}{c|}{\it Spanish Base} & \multicolumn{4}{c|}{\it Multilingual Base} & \multicolumn{2}{c}{\it Large}\\
Dataset & Beto & Bertin & Elect. & MarIA & IXAes & IXAm	& mBERT & XLM-R & mDeB3 & MarIA {\bf L} & XLM-R{\bf L} \\
\midrule
PoS UD  & 99.00 &	98.98 &	98.18 &	\underline{99.07} &	99.03 &	98.90 &	99.01 & 99.02 &	\underline{99.05} & 99.04 &	\underline{\bf 99.11} \\
PoS Capitel &  98.36 & 98.47 & 98.16 & 98.46 & \underline{98.55} & 98.32 & 98.39 & 98.47 & \underline{98.56} & 98.56 &	\underline{\bf 98.63} \\
NERC CoNLL & 87.59	& 88.35 & 79.54 & 88.51	& \underline{88.70}	& 87.85	& 86.91 & 88.11 & \underline{88.73}	& 88.23	& \underline{\bf 89.02} \\
NERC Ancora & 92.46 & 92.15	& 85.66	& 93.34 & \underline{\bf 93.57}	& 92.58	& 92.58	& 92.47	& \underline{93.02}	& 92.45 & \underline{93.13} \\
NERC Capitel & 87.72 & 88.56 & 80.35 & 89.60 & \underline{89.83} & 88.65 & 88.10 & 88.55 & \underline{89.86} & \underline{\bf 90.51} & 90.19 \\
STS  &  81.59 &	79.45 &	80.63 &	\underline{\bf 85.33} &	83.82 &	83.09 &	81.64 & 83.47 & \underline{83.61} & \underline{84.11} &	84.04 \\
MLDoc  &  \underline{\bf 97.14} & 96.68 & 95.65 & 96.64	& 96.78	& \underline{96.70}	& 96.17	& 96.30	& 96.62 & 97.02	& \underline{97.05} \\
PAWS-X  &  89.30 & 89.65 & \underline{90.45} & 90.20 & 89.99 & 88.06 & 90.00 & 89.82 & \underline{91.90} & 91.50 & \underline{\bf 91.93} \\
XNLI  &  81.30 & 78.90 & 78.78 & 80.16 & \underline{82.40} & 79.40	& 78.76	& 81.14	& \underline{84.85} & 82.63	& \underline{\bf 84.95} \\
SQAC  &  \underline{79.23} & 76.78 & 73.83 & \underline{79.23} & 78.91 & 77.38 & 75.62 & 77.28 & \underline{80.78} & 82.02 & \underline{\bf 84.10} \\
CoMeta  & 64.28 & 61.52	& 61.18	& 63.08	& \underline{64.79}	& 62.04	& 61.77	& 63.82 & \underline{\bf 67.46} & 62.02	& \underline{67.44} \\
\midrule
Average  &  87.09 & 86.32 & 83.86 & 87.60 & \underline{87.85} & 86.63 & 86.27 & 87.13 & \underline{88.59} & 88.01 & \underline{\bf 89.05} \\
Average$^*$ & 89.37 & 88.80	& 86.12	& 90.05	& \underline{90.16}	& 89.09	& 88.72 & 89.46	& \underline{90.70}	& 90.61	& \underline{\bf 91.22} \\
Wins group & 1.5 &  &  1 & 2.5 & \underline{6}     &    1     &  &  & \underline{10} &  2  & \underline{9}\\ 
Wins all   & 1 &  &      & 1 & 1  &          &  &      & 1 &   1 & \textbf{6}\\
\bottomrule
\end{tabular}
}
\caption{Results with models grouped according to: Spanish
base-size, multilingual base-size, and large-size (one Spanish and one
multilingual). Best result per group with \underline{underline}, best result
overall in \textbf{bold}. We report average across datasets, average$^*$
without the metaphor dataset CoMeta, wins in each group
and wins overall (ties are scored as $1/n$ where $n$ is systems tied). Metric
F1 micro except for MLDoc and XNLI (accuracy); STS is evaluated on the
official \emph{combined score}. For space reasons we only report results from
one Ixa monolingual model: IXAes $=$ IXABERTesv2.}
\label{tab:result}
\end{center}
\end{table*}

\section{Experimental setup}\label{sec:experimental-setup}

Our experimental setup follows the one proposed by MarIA
\cite{gutierrezfandino2022}, with the caveat that we include 6 more language
models in our evaluation and two extra datasets. Thus, the 12 models listed in
Table \ref{tab:spanish-lm} are evaluated on 8 tasks and 11 datasets: For POS tagging the
UD and Capitel datasets \cite{Taul2008AnCoraMA,Zamorano2020OverviewOC}; for NER we use CoNLL 2002
\cite{Sang2002IntroductionTT}, Capitel \cite{Zamorano2020OverviewOC} and Ancora
2.0 \cite{Taul2008AnCoraMA}; the
Semantic Text Similarity dataset is based on the data by 
\namecite{Agirre2014SemEval2014T1} and \namecite{Agirre2015SemEval2015T2};
MLDoc \cite{Schwenk2018ACF} for document classification; paraphrase
identification with PAWS-X \cite{Yang2019PAWSXAC}, XNLI for Natural Language
Inference \cite{Conneau2018XNLIEC}, Question Answering with the SQAC data
\cite{gutierrezfandino2022} and CoMeta \cite{SanchezBayona2022LeveragingAN} for
metaphor detection.

For comparison purposes, we use the same data splits as in the MarIA paper. For the
two datasets added for this paper, Ancora 2.0 NER and CoMeta, we make public
the splits we created. Both Ancora 2.0 and CoMeta are publicly available and we
thought that they were a good addition to the benchmark. In this sense, it
should be noted that every dataset is public except the Capitel POS and NER
corpora.
We are not particularly
fond of using data which is not publicly available, at least for research,
because it makes reproducibility impossible thereby hindering the progress of
scientific research. However, we decided to include them to make it a more
comprehensive comparison with previous work on benchmarking language models in
Spanish.

For fine-tuning the models we use the same scripts used by \namecite{gutierrezfandino2022} as available in their
Github repository\footnote{\url{https://github.com/PlanTL-GOB-ES/lm-spanish}}
with minor modifications. For
each task, a single linear layer is added on top of the model being fine-tuned.
In the case of sentence/paragraph-level classification tasks, the
\texttt{[CLS]} token is used for BERT models, and the
\texttt{{\textless}s\textgreater} token in the case of RoBERTa models. 
We use maximum sequence length of 512.  A grid search of
hyperparameters is performed to pick the best batch size (8, 16, 32), weight decay (0.01,
0.1) and learning rate (1e-5, 2e-5, 3e-5, 5e-5). We pick the best model on the
development set over 5 epochs. We keep a fixed seed to ensure reproducibility
of results. The experiments have been implemented using the HuggingFace
Transformers API \cite{Wolf2019TransformersSN}. Code and data splits are publicly
available\footnote{\url{https://github.com/ragerri/evaluation-spanish-language-models}}.

\section{Results}\label{sec:results}

Table \ref{tab:result} shows the results for each model in each dataset.
Results already reported by \namecite{gutierrezfandino2022} are included here
verbatim. The rest of the results have been obtained by fine-tuning the models
following the method described in the previous section. The
average across datasets and the number of datasets where one method wins over
the rest allow to set a clear picture.

First, among Spanish-only base models, the best results are obtained by IXAes,
which performs better than MarIA (the second best) in both average and wins in
datasets. They are followed by BETO, BERTIN and finally Electricity. This
result is interesting as IXAes is trained with a much smaller public corpus.

Second, if we look at the multilingual base models, mDeBERTa is the clear
winner, followed by XLM-RoBERTa and ixambert which perform quite similarly.
 
Third, if we compare monolingual and multilingual base models, the monolingual
IXAes outperforms the best comparable multilingual model, XLM-RoBERTa. However, the
newer mDeBERTa yields the best results overall. It should be noted that all the
Spanish models were produced before the DeBERTa v3 architecture was introduced,
which may perhaps explain their lower results. 
    
Fourth, regarding the largest models, XLM-RoBERTa outperforms MarIA large in 9
out of 11 datasets, and obtains a better average performance. In fact, even
mDeBERTa obtains slightly better results than MarIA large. Moreover, the pre-existing XLM-RoBERTa model works for 99
additional languages, allowing also to perform cross-lingual transfer. The only
single disadvantage is that the size of XLM-RoBERTa is larger, mostly due to
its larger vocabulary size, but the cost in running time (Flops) is comparable
for both. 
    
Overall, results demonstrate that XLM-RoBERTa-large is the best model across
the board, including the newer mDeBERTa. The
DeBERTa team have not reported results or released a large DeBERTa multilingual
model, but given the strong results of the English DeBERTa large model
\cite{He2021DeBERTaV3ID}, it can be assumed that its results may be superior to
those obtained by XLM-RoBERTa-large.

Finally, it should be noted that for the task of metaphor detection the results
are significantly lower across the board. This is not entirely surprising, as
the state-of-the-art in metaphor detection is in general quite low. In any
case, and motivated by this fact, we also calculated the average$^*$ without
taking into account the metaphor detection results. As it can be seen, while
the results get slightly higher, the trends discussed still hold.

\section{Discussion}\label{sec:discussion}

According to the results, the following lessons can be drawn.

\paragraph{Which model should I use according to my computing budget?} If the
user is interested in best results at inference, XLM-RoBERTa-large is nowadays the best
option, at the cost of requiring more time and GPU memory. mDeBERTa 
would be the next best choice for smaller memory and runtime budgets. For a
more modest solution, IXAes would be a good choice.

\paragraph{Which model should I use according to my task?} In this work we
cover a broad but limited number of datasets. If your target task is similar to
one of the datasets, then you might want to use the model that excels at this
task and that meets your budget requirements (in terms of the GPU hardware that
it can be afforded). For most tasks XML-RoBERTa-large
is the best option, with the additional benefit from cross-lingual transfer. 
For smaller budgets we recommend to check the
underlined results in the different groups in Table \ref{tab:result}. For the
cases where your target task is not covered, the safest option is to take the
best overall model according to your budget.

\paragraph{Is there an explanation for the lower performance of some models?}
Larger models are expected to perform better. Furthermore, the mDeBERTa results
are not particularly surprising. However, in the case of models with the same
architecture and size, it would be good to be able to pinpoint the causes for
the disappointing performance of some models. 

An important factor could be the \textbf{corpora} used. In principle the MarIA
models use the largest and, according to their authors, the cleanest corpus for
Spanish ever produced. However, it turns out that, for the same base size,
IXAes gets better results, even if it was trained on a smaller corpus (OSCAR)
which is publicly available since 2019 \cite{OrtizSuarezSagotRomary2019}. OSCAR
is based on Common Crawl, covers 166 languages, and uses a very light publicly
available filtering software, while the BNE corpus was filtered in-house
following previous work \cite{Virtanen2019MultilingualIN}. The strongest
performers (XLM-RoBERTa and mDeBERTa) also use a filtered version of Common
Crawl, CC100, which in this case was publicly released by Facebook around 2020
\cite{Conneau2020UnsupervisedCR}. There are evidences that high-quality
filtering does not improve downstream performance and that size seems to be
equally important \cite{Artetxe2022DoesCQ}. Perhaps an audit of a sample of the
BNE corpus compared with the other corpora used to train the models would
provide further light on this issue. On this line of research, two possible strategies would be to:
(i) use the same architecture and training procedure but with different corpora
\cite{Artetxe2022DoesCQ}; (ii) fix the corpus used for training varying the
training method and specifications.

Other explanations may be related to how much training procedure and
hyperparameters vary from one model to the other (see Section
\ref{sec:spanish-language-models}). Although an exhaustive analysis is not
feasible, two key factors could be the \emph{size of the vocabulary}
\cite{zheng-etal-2021-allocating} and the number of \emph{training examples
seen in training}. In fact, the Spanish models have relatively small
vocabularies compared to their XLM-RoBERTa and DeBERTa counterparts, and BETO and
Electricidad have smaller vocabulary size than the better performing IXAes and
MarIA. Thus, vocabulary size might be part of the explanation, but it does not
explain the differences in results between the Spanish models with the same
vocabulary, so we may need to consider other possible explanations.

If we look at the number of steps in training, MarIA uses a strategy which is
substantially different to the rest of the models, in particular to XLM-RoBERTa
and mDeBERTa. Both longer \cite{Devlin19,Conneau2020UnsupervisedCR} and shorter
\cite{Komatsuzaki2019OneEI} training have been recommended. In the light of the results, one would say
that the strategy from XLM-RoBERTa and mDeBERTa is the best, so in this case it
would look like as if some of the Spanish models have been undertrained.
However, in order to have a more conclusive answer, it would be necessary to
experiment with the number of steps fixing the other variables involved in the
training process.

Summarizing, it seems that publicly available corpora suffice for optimal
results, and that the larger the model and the vocabulary the better.
Additionally, the number of steps could also play an important role.
Unfortunately, the post-hoc analysis carried out in this paper cannot give a
more precise picture, and carefully designed experiments along the lines of
the ones suggested above would be necessary to shed some more light and perhaps
to improve results.

\paragraph{Training a monolingual model, is it worth it?}  Common wisdom
indicates that monolingual models improve over multilingual models
\cite{martin-etal-2020-camembert,agerri-etal-2020-give,Virtanen2019MultilingualIN,Tanvir2021EstBERTAP,armengol-estape-etal-2021-multilingual},
which led to a proliferation of models for many target languages. Most of the
models have been shown to outperform their multilingual counterparts, but often
have only considered multilingual BERT completely ignoring XLM-RoBERTa \cite{Nozza2020WhatT}. 

Part of the mixed signals could be also caused by the size of the language:
while large languages like Spanish and English are very well represented in
multilingual models, low-resource languages tend to have a very small quota of
training instances. Training a model using larger amounts of better quality corpora for
low-resource languages could thus explain the good results
of monolingual models with respect to multilingual ones \cite{agerri-etal-2020-give,Bhattacharjee2021BanglaBERTCE,Nzeyimana2022KinyaBERTAM}, but this may not
necessarily be the case for high-resource languages, as evidenced by the
results reported in Table \ref{tab:result}.

Our work shows that some monolingual base models such as IXAes or MarIA do
slightly improve over the results of a comparable XLM-RoBERTa-base multilingual
model. However, the two best performing models for Spanish are currently
mDeBERTa (base) and XLM-RoBERTa-large. Considering these results and
the literature mentioned above, it would seem that the amount and quality of
publicly available Spanish corpora suffices, and that future improvements will
need to come from larger models or architecture improvements, as shown by
DeBERTa or T5 for English, or by careful experimentation as outlined above.

\begin{table*}
\begin{center}
\resizebox{1\linewidth}{!}{
\begin{tabular}[b]{l|@{\hspace{0.2cm}}c@{\hspace{0.2cm}}c@{\hspace{0.2cm}}c@{\hspace{0.2cm}}c@{\hspace{0.2cm}}c@{\hspace{0.2cm}}|@{\hspace{0.0cm}}c@{\hspace{0.2cm}}c@{\hspace{0.2cm}}c@{\hspace{0.2cm}}c|@{\hspace{0.0cm}}c@{\hspace{0.2cm}}c@{\hspace{0.2cm}}|c}
\toprule
& \multicolumn{5}{c|}{\it Spanish Base} & \multicolumn{4}{c|}{\it Multilingual Base} & \multicolumn{2}{c}{\it Large} & \multicolumn{1}{|c}{Prev} \\
Dataset & Beto & Bertin & Elect. & MarIA & IXAes & IXAm	& mBERT & XLM-R & mDeB3 & MarIA {\bf L} & XLM-R{\bf L} & SOTA \\
\midrule
PoS UD  & 99.10	& 99.11	& 98.37	& 99.14	& \underline{99.17}	& 98.98	& 99.01	& 99.16	& \underline{\bf 99.20} & 99.12	& \underline{99.19} & 99.05 \\
PoS Capitel &  98.57 & 98.63 & 98.40 & 98.67 & \underline{98.75} & 98.55 & 98.60 & 98.68 & \underline{98.76} & 98.73 & \underline{\bf98.82} & - \\
NERC CoNLL & 87.59	& 88.35 & 79.54 & 88.51	& \underline{88.70}	& 87.85	& 86.91 & 88.11 & \underline{88.73}	& 88.23	& \underline{\bf 89.02} & 95.90 \\
NERC Ancora & 92.46 & 92.15	& 85.66	& 93.34 & \underline{\bf 93.57}	& 92.58	& 92.58	& 92.47	& \underline{93.02}	& 92.45 & \underline{93.13} & - \\
NERC Capitel & 87.72 & 88.56 & 80.35 & 89.60 & \underline{89.83} & 88.65 & 88.10 & 88.55 & \underline{89.86} & \underline{\bf 90.51} & 90.19 & 90.34 \\
STS  &  81.59 &	79.45 &	80.63 &	\underline{\bf 85.33} &	83.82 &	83.09 &	81.64 & 83.47 & \underline{83.61} & \underline{84.11} &	84.04 & - \\
MLDoc  &  \underline{\bf 97.14} & 96.68 & 95.65 & 96.64	& 96.78	& \underline{96.70}	& 96.17	& 96.30	& 96.62 & 97.02	& \underline{97.05} & 96.80 \\
PAWS-X  & 89.15 & 90.35 & 89.20 & 90.45 & \underline{90.75} & 89.15 & 89.30 & 90.35 & \underline{\bf 92.50} & 90.95 & \underline{92.05} & 90.70 \\
XNLI  &  81.30 & 78.90 & 78.78 & 80.16 & \underline{82.40} & 79.40	& 78.76	& 81.14	& \underline{84.85} & 82.63	& \underline{\bf 84.95} & 85.50 \\
SQAC  &  \underline{79.23} & 76.78 & 73.83 & \underline{79.23} & 78.91 & 77.38 & 75.62 & 77.28 & \underline{80.78} & 82.02 & \underline{\bf 84.10} & - \\
CoMeta  & 64.28 & 61.52	& 61.18	& 63.08	& \underline{64.79}	& 62.04	& 61.77	& 63.82 & \underline{\bf 67.46} & 62.02	& \underline{67.44} & 67.46\\
\midrule
Average &  87.10 & 86.41 & 83.78 & 87.65 & \underline{87.95} & 86.76 & 86.22 & 87.21 & \underline{88.67} & 87.98 & \underline{\bf 89.09} \\
Average$^*$ & 89.39 & 88.90 & 86.04 & 90.11 & 90.27 & 89.23 & 88.67 & 89.55 & 90.79 & 90.58 & 91.25 \\
Wins group & 1.5 &  &  & 1.5 & \underline{8}     &    1     &  &  & \underline{10} &  2  & \underline{9}\\ 
Wins all   & 1 &  &      & 1 & 1  &          &  &      & 3 &   1 & \textbf{4}\\
\bottomrule
\end{tabular}
}
\caption{Same results as in Table 2, but using standard metrics (accuracy for
PAWS-X, word accuracy for PoS UD and Capitel). We also report previous
state-of-the-art results where available. See text for details.}
\label{tab:mod-results}
\end{center}
\end{table*}

\paragraph{Better research reporting practices should be encouraged.} The
XLM-RoBERTa models were widely known and available when the Spanish models were built, but none of the
publications on language models in Spanish compared their results to
XLM-RoBERTa, implicitly sending the wrong message that ignoring XLM-RoBERTa was the best option 
when working with Spanish language models. As our results show,
XLM-RoBERTa is currently the strongest option to build NLP applications in Spanish.

\paragraph{Comparison to the state-of-the-art.} In relation to the previous
point, research on language models seem to be inadvertently forgetting the
primary objective of building language models in the first place, namely,
improving the state-of-the-art of NLP technology. Thus, previous published work
do not mention what the state-of-the-art is
for each of the tasks used to benchmark the models. Without doing so, it is
just not possible to know how much a given language model is
actually advancing NLP technology. Therefore, we first 
reevaluate three tasks (PAWS-X and Capitel and UD POS) to report the most
common accuracy metric usually used for those tasks (instead of the F1 score
used in previous evaluations of language models in Spanish). Table \ref{tab:mod-results} offers the
overall results with PAWS-X, Capitel and UD PoS evaluated using accuracy. The
new results were obtained by fine-tuning all 12 models following the methodology provided
in Section \ref{sec:experimental-setup}. As it can be seen, they confirm the
trends already observed and discussed above.

Based on Table  \ref{tab:mod-results} we can now compare the results of the
models with respect to the state-of-the-art in each task. First, it should be
noted that for five tasks (Capitel PoS, Ancora 2.0 NER, STS, SQAC and CoMeta) 
their results have been published for the first time during the evaluation of language models
in Spanish (including this one). Out of the six remaining tasks, the best
results of the models on NERC CoNLL and XNLI remain far from the state-of-the-art reported by
\namecite{wang-etal-2021-automated} and \namecite{Aghajanyan2021BetterFB}, with
a 95.90 F1 score for NERC and 85.50 in accuracy in XNLI. For PoS UD, our best
model scores 99.20 (mDeBERTa), comparable to 
\cite{Straka2019EvaluatingCE}, which scored 99.05. The same can be said
regarding NERC Capitel, where the difference between the best score by MarIA
large (90.51) and the previous best (90.34) is rather
anecdotical \cite{Agerri2020ProjectingHA}, and MLDoc, for which BETO slightly
outscores 97.17 vs 96.80, the previous best result published \cite{Lai2019BridgingTD}.
Finally, for PAWS-X only XLM-RoBERTa and mDEBERTa clearly outperform the
state-of-the-art previously reported by \namecite{Yang2019PAWSXAC}.

Summarizing, out of the 11 datasets, the Spanish monolingual language models obtain
minimal better results for three tasks only: PoS UD, NERC Capitel
and MLDoc, although the differences are too small to be significant.
Furthermore, they underperform in the rest of the tasks with respect to previously published
state-of-the-art results.

\paragraph{What should be the next steps for Spanish models?} One could argue
that given the better results of the multilingual models released by large
companies, there is no need to devote resources to build better models for
Spanish. Unfortunately, there is no guarantee that large companies will keep
releasing updated models, which will make the models obsolete very quickly. As
an example, all models are trained on texts before Covid-19, and thus have no
notion of what the latest pandemic is about. It will also leave the leadership
of NLP for Spanish at the hand of third parties. Given the foundational nature
of language models it is necessary to ensure that new updated versions of the
best performance are produced regularly. 

Our analysis has shown that it is not trivial to produce high-performance language models, as it is still
an open, resource-heavy, research problem. In addition, new and powerful models
are being developed at a fast pace, including encoder-decoder models like T5
\cite{t5}, with its superior performance in many downstream tasks when compared to
encoder-only models like BERT \cite{Devlin19}, or decoder-only models like
GPT-3 \cite{gpt3}, which has facilitated good results in generation tasks, but
also in zero- and few-shot approaches to regular NLP tasks
\cite{gpt3}.

In other countries other than Spain, policy-makers and research funding
agencies have recognised the strategic importance of this field and its
research-intensive and ambitious nature. For example, the European Language
Equality (ELE) project\footnote{\url{https://european-language-equality.eu}}
has defined an European strategy where three main requirements are identified:
expert researchers, (public) data, and computational power (GPUs). However,
expert researchers with experience in this field do not abound, and the GPUs
needed are a substantial investment which should be carefully designed to meet
the demands of training language models. 

In our opinion, it is necessary to launch a multi-year research program devoted to
language models in Spanish, which should match the ambition of this strategic field and
which should marry the following: (i) The expertise of the best researchers in
the field of language models. 
Unfortunately they are a scarce resource, as they are being
actively recruited by large companies. We believe that only an attractive research landscape
which includes the resources mentioned next will allow to attract them to this
program. (ii) The necessary resources, either monetary or in the form of
sustained access to powerful GPUs. In order to explore and understand the
reasons for the results reported here, it is necessary to set an experimental program where variants of language
models are trained on different experimental conditions.

\section{Conclusion}


In this paper we have presented a comprehensive head-to-head comparison of
language models for Spanish. The results show that (i) multilingual models from
large companies fare better than monolingual models; (ii) results across the
monolingual Spanish models are not conclusive, with supposedly smaller and
inferior models performing competitively. Based on these empirical results, we
have argued for the need of further research to understand the factors
underlying these results. Thus, the effect of corpus size, quality and
pre-training techniques need to be further investigated to be able to obtain
Spanish monolingual models significantly better than the multilingual ones
released by large private companies, specially in the face of rapid ongoing
progress in the field. 

While the recent activity in the development of language technology for
Spanish is to be welcomed, our results show that building language models
remains an open, resource-heavy problem which requires to marry monetary and
computational resources with the best research expertise and practice.

Other future work should include GPT-3 style improvements at scale for Spanish.
Furthermore, most of the current few-shot and generative-related work for
languages other than English is being done with multilingual models such as
mBART and mT5. Thus, a lot of work remains to be done if Spanish as language is to be at
the forefront of language technology.

\section*{Acknowledgments}

We would like to thank the authors of MarIA models for their valuable help in using
their evaluation scripts. This has allow us to follow the same evaluation
methodology thereby facilitating comparability of results.

This work has been partially supported by the HiTZ center and the Basque
Government (Research group funding IT-1805-22). We also acknowledge the funding
from the following projects: (i) DeepKnowledge (PID2021-127777OB-C21)
MCIN/AEI/10.13039/501100011033 and ERDF A way of making Europe; (ii) Disargue
(TED2021-130810B-C21), MCIN/AEI/10.13039/501100011033 and European Union
NextGenerationEU/PRTR (iii) Antidote (PCI2020-120717-2),
MCIN/AEI/10.13039/501100011033 and by European Union NextGenerationEU/PRTR;
(iv) DeepR3 (TED2021-130295B-C31) by MCIN/AEI/10.13039/501100011033 and
EU NextGeneration programme EU/PRTR.  Rodrigo Agerri currently holds the
RYC-2017-23647 fellowship (MCIN/AEI/10.13039/501100011033 and by ESF Investing
in your future).

\bibliographystyle{fullname}

\bibliography{references}

\end{document}